\documentclass[12pt]{article}

\usepackage{amsmath, amssymb, graphicx, cite, hyperref}
\usepackage[margin=1in]{geometry}
\usepackage[utf8]{inputenc}
\usepackage{graphicx}
\usepackage{float}  
\usepackage{amsmath}
\usepackage{booktabs}
\usepackage{placeins}  
\usepackage{plantuml}
\usepackage{geometry}  
\usepackage{hyperref}  
\usepackage{subcaption} 
\usepackage{array}
\usepackage{multirow}

\title{Precision Adaptive Imputation Network : An Unified Technique for Mixed Datasets}

\author{
    Harsh Joshi$^{1}$, Rajeshwari Mistri$^{1}$, Manasi Mali$^{1}$, \\
    Parul Kumari$^{1}$, Nachiket Kapure$^{1}$ \\[1.5ex]
    $^1$B.K. Birla College of Arts, Science and Commerce, Kalyan \\[1.5ex]
    \texttt{\{joshiharsh0506, rajeshwarimistri11, malimanasi2002,} \\
    \texttt{parulkumari2307, kapnachi1904\}@gmail.com}
}

\date{}

\begin{document}

\maketitle
    \begin{abstract}
The challenge of missing data remains a significant obstacle across various scientific domains, necessitating the development of advanced imputation techniques that can effectively address complex missingness patterns. This study introduces the Precision Adaptive Imputation Network (PAIN), a novel algorithm designed to enhance data reconstruction by dynamically adapting to diverse data types, distributions, and missingness mechanisms. PAIN employs a tri-step process that integrates statistical methods, random forests, and autoencoders, ensuring balanced accuracy and efficiency in imputation. Through rigorous evaluation across multiple datasets, including those characterized by high-dimensional and correlated features, PAIN consistently outperforms traditional imputation methods, such as mean and median imputation, as well as other advanced techniques like MissForest. The findings highlight PAIN's superior ability to preserve data distributions and maintain analytical integrity, particularly in complex scenarios where missingness is not completely at random. This research not only contributes to a deeper understanding of missing data reconstruction but also provides a critical framework for future methodological innovations in data science and machine learning, paving the way for more effective handling of mixed-type datasets in real-world applications. 

\noindent{\textbf{Keywords:}}
Data Imputation, Missing Data, MNAR, PAIN, Machine Learning, Statistical Methods, Random Forests 
\end{abstract} 

\section{Introduction}
In the era of big data, where decisions in healthcare and finance can be life-altering, incomplete data is a critical liability. Imagine a clinical trial where patient histories are partially missing due to privacy restrictions or consider a financial model undermined by missing customer data during a credit risk assessment. These gaps not only threaten model reliability but also lead to costly, and sometimes catastrophic, consequences. Addressing this challenge requires robust, adaptable imputation techniques that can cope with the messy, mixed, and incomplete nature of real-world data.

\medbreak

Albeit significant advancements, existing imputation techniques fall short in addressing key challenges. Many methods excel under the assumption of data Missing Completely at Random (MCAR), yet they struggle with scenarios involving Missing at Random (MAR) or Missing Not at Random (MNAR), where missingness depends on unobserved variables. Although Neural network-based approaches, are promising, they often rely on restrictive assumptions about data distribution and require extensive hyperparameter tuning, limiting their practical applicability. Furthermore, current methods frequently fail to accommodate mixed-type datasets—those containing a combination of numerical, categorical, and ordinal variables. For instance, healthcare datasets often include patient demographics (categorical), medical histories (ordinal), and lab results (numerical), yet most imputation tools are optimized for a single data type. These gaps compel practitioners to rely on multiple tools or compromise on suboptimal results.

\medbreak

This study introduces the Precision Adaptive Imputation Network (PAIN), a novel algorithm is designed to address the limitations of existing techniques. PAIN adapts dynamically to varying data types, distributions, and missingness patterns, ensuring robust imputation in diverse scenarios. Its tri-step process includes: (1) a hybrid architecture that integrates statistical methods, random forests, and autoencoders for balanced accuracy and efficiency; (2) an adaptive weighting mechanism that tailors imputation strategies based on data characteristics; and (3) a refinement layer that polishes imputed data, reducing noise and preserving integrity. By unifying these elements, PAIN delivers superior imputation performance.

\medbreak

This paper is organized to systematically demonstrate the development and impact of PAIN. Section 2, reviews the existing data imputation techniques, highlighting their limitations and setting the stage for adaptive framework. In Section 3, the methodology behind PAIN is detailed, while Section 4 presents a rigorous evaluation across diverse datasets and missingness scenarios. Following this, the study in Section 5 discusses the implications of our findings, offering insights for both researchers and practitioners. Finally, Section 6 concludes by summarizing the contributions and exploring future directions for adaptive imputation.

\section{Literature Review}
The persistent challenge of missing data continues to plague researchers across diverse
scientific domains, from healthcare and genomics to environmental sciences and financial
modeling \cite{article,Afkanpour2024-dq,Alwateer2024-pg}. Modern datasets increasingly demonstrate complex missingness patterns that
challenge traditional imputation approaches, necessitating more sophisticated methodological
interventions.

\medbreak

The evolutionary trajectory of data imputation techniques reveals a remarkable transformation from rudimentary statistical methods to advanced machine learning and hybrid approaches. Initially, researchers relied on deterministic techniques like mean substitution, median replacement, and basic k-Nearest Neighbors (kNN) algorithms \cite{Alwateer2024-pg,Khan2020-uo}. These traditional approaches demonstrated significant limitations, particularly when confronted with missingness rates exceeding 20\%, revealing critical gaps in handling complex data structures \cite{Khan2020-uo,Schreiber2023-rv}.

\medbreak

While, Machine Learning (ML) and Deep Learning (DL) techniques marked a revolutionary paradigm shift in addressing missing data challenges. Probabilistic methods such as Multiple Imputation by Chained Equations (MICE) and innovative approaches like Nearest-Neighbor Kernel Density Estimation (NNKDE) emerged as powerful alternatives \cite{Schreiber2023-rv,Seu2022-ag, Lalande2023-gl}. These advanced techniques demonstrated superior capabilities in preserving intricate data distributions and capturing nuanced inter-variable relationships across multiple domains. Generative Adversarial Networks (GANs) emerged as a particularly promising imputation technique, especially for handling MAR and MNAR scenarios \cite{Hameed2023-er,Ismail2022-uk}. Empirical studies have highlighted GAN’s remarkable potential in managing datasets with missingness rates up to 50\%, particularly in complex domains like healthcare and electronic health records \cite{Afkanpour2024-dq,Hameed2023-er}. However, these advanced methods are not without challenges, with computational intensity and training stability remaining significant limitations.

\medbreak

The landscape of imputation techniques has witnessed remarkable innovations. The Denoising Self-Attention Network (DSAN) demonstrated exceptional performance in handling both numerical and categorical data \cite{Lee2023-bu}, while reinforcement learning models introduced dynamic strategies for optimizing imputation accuracy \cite{Awan2022-tt}. A groundbreaking approach, the Contextual Large Language Model for Imputation (CLAIM), leveraged pre-trained language models to provide contextually adaptive imputation strategies \cite{Hayat2024-kr}, representing a significant leap in addressing missing data challenges. 

\medbreak

Domain-specific research has unveiled nuanced insights into imputation effectiveness. While, genomic studies consistently demonstrated that kNN and Random Forest techniques outperformed traditional methods, maintaining accuracy even under extreme missingness scenarios \cite{Berdnikova2024-yc,Barrabes2024-oz}. Specialized techniques like Makima interpolation proved particularly effective in specific domains such as smart grid data analysis \cite{Schreiber2023-rv}, highlighting the importance of context-specific imputation strategies.

\medbreak

The computational landscape of imputation techniques reveals both promise and challenges, as Multilevel stochastic optimization approaches for medical datasets \cite{Li2024-sd} and in-database imputation techniques \cite{Perini2024-uw} represent innovative solutions to computational constraints. These methods demonstrate the potential for more efficient imputation strategies, particularly for large-scale datasets with complex missingness patterns.

\medbreak

Comparative analyses across multiple studies consistently revealed the superiority of advanced imputation techniques. While traditional methods struggled with complex data patterns, the ML approaches showed remarkable adaptability. The Enhanced GAN approach \cite{Qin2024-wa} addressed critical challenges such as gradient vanishing and mode collapse, offering improved imputation accuracy and runtime efficiency. Although significant advancements persist critical research gaps, where computational efficiency remains a primary challenge, with many advanced techniques requiring substantial computational resources. The field lacks standardized benchmarking frameworks across different datasets and domains, hindering comprehensive comparative analysis. Moreover, most research has predominantly focused on MCAR and MAR scenarios, leaving MNAR mechanisms relatively unexplored \cite{Alabadla2022-tt,Umar2023-ox}.

\medbreak

Future research opportunities are abundant. There is a pressing need for hybrid models that can effectively combine probabilistic and ML techniques, addressing both linear and non-linear data dependencies. Domain-specific adaptations, particularly for mixed data types and real-world MNAR scenarios, represent critical areas for further investigation. The emerging field of contextual and probabilistic imputation techniques, such as the CLAIM approach utilizing large language models \cite{Hayat2024-kr} and probabilistic nearest-neighbor methods \cite{Dinh2021-fe}, demonstrates the potential for more adaptive and intelligent imputation strategies. These approaches offer promising directions for addressing the complex challenges of missing data across diverse domains.

\section{Methodology}
The methodology encompasses experimental design, implementation of the proposed framework, and comparative evaluation against established imputation methods. The core innovation of this research lies in the PAIN three-layer architecture.
\medbreak
\begin{figure}[h!]
    \centering
    \includegraphics[width=0.8\textwidth]{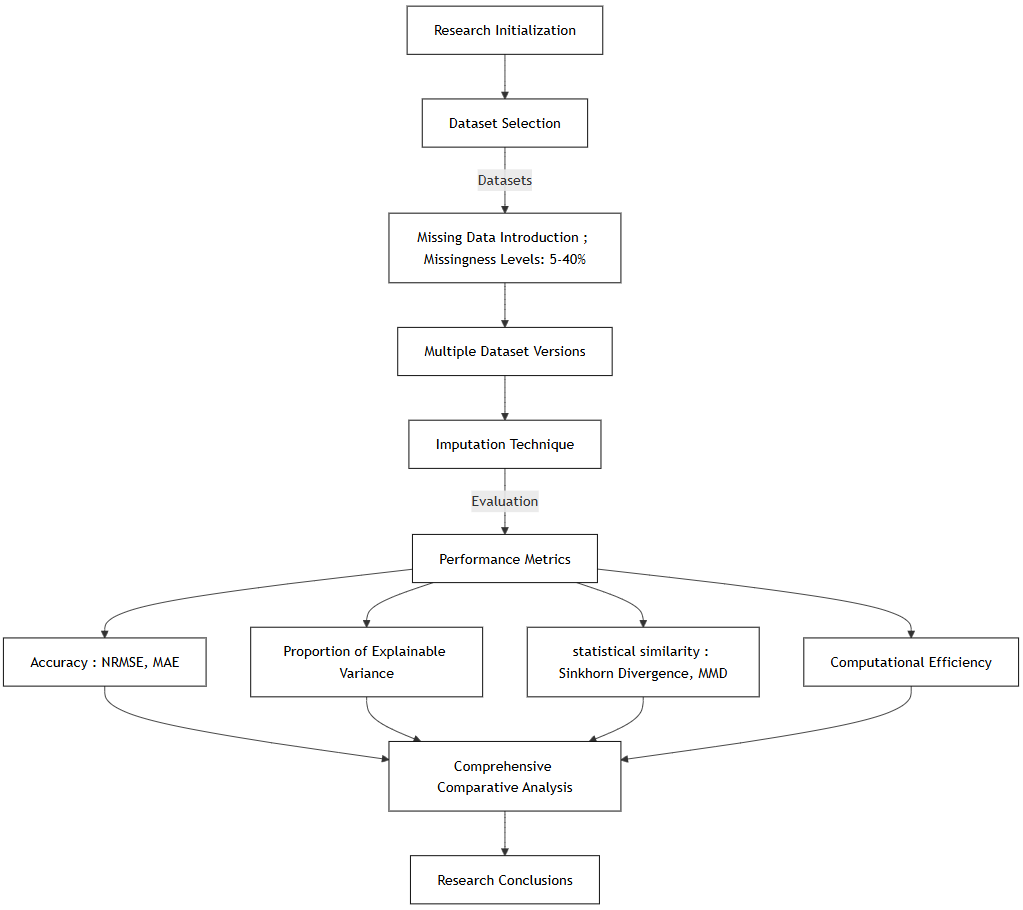}
    \caption{Illustrates the Research Methodology, highlighting the Pipeline from Dataset Preparation to Comparative Analysis.}
    \label{fig:methodology}
\end{figure}

\medbreak
\vspace{10pt} 
\textbf{Experimental Setup}

This study evaluates imputation methods and introduces PAIN, a multi-level framework for managing missing data. The research is based on an experimental setup using open-source datasets, is designed to systematically assess the performance of imputation techniques across different data structures and missingness mechanisms (see figure \ref{fig:methodology}). The dataset is used from UCI machine learning dataset, as Table \ref{tab:Resource Utilization Metrics}. gives a snapshot of information about the dataset being utilized.

\begin{table}[h!]
\centering
\renewcommand{\arraystretch}{1.2}
\caption{Resource Utilization Metrics}
\adjustbox{max width=\textwidth}{%
\begin{tabular}{@{}p{2.5cm}>{\centering\arraybackslash}p{2.5cm}>{\centering\arraybackslash}p{2.5cm}>{\centering\arraybackslash}p{2.5cm}>{\centering\arraybackslash}p{2.5cm}>{\centering\arraybackslash}p{2.5cm}@{}}
    \toprule
    \textbf{Dataset Name}       & \textbf{\# Columns} & \textbf{\# Rows} & \textbf{Continuous Columns} & \textbf{Discrete Columns} & \textbf{Normally Distributed Columns} \\
    \midrule
    Maternal Health Risk \cite{Ahmed2020-og}   & 7                   & 1013             & 5                            & 2                         & 0 \\
    Concrete Data \cite{Yeh1998-yt}          & 9                   & 1030             & 9                            & 0                         & 0 \\
    Absenteeism \cite{Andrea_Martiniano2012-ub}            & 20                  & 740              & 8                            & 12                        & 0 \\
    Forest Fire \cite{Paulo_Cortez2007-re}            & 12                  & 517              & 4                            & 9                         & 0 \\
    Wine \cite{Stefan_Aeberhard1992-ff}                   & 12                  & 1599             & 1                            & 11                        & 0 \\
    \bottomrule
\end{tabular}
}
\label{tab:Resource Utilization Metrics}
\end{table}

Table \ref{tab:Resource Utilization Metrics}. Nature of datasetFurthermore, missing data was introduced using MAR patterns at five levels (5\%, 10\%, 20\%, 30\%, 40\%). For each level, five independent datasets are generated to ensure statistical validity. This setup allows a balanced evaluation of imputation methods across diverse conditions.

\textbf{Imputation Methods}

A wide range of imputation methods are evaluated, including statistical, ML, and neural network-based techniques. Custom algorithms and the proposed PAIN are included for comparative analysis (see Table \ref{tab:Categorization of Techniques in Data Processing}).

\begin{table}[h]
    \centering
    \caption{Categorization of Techniques in Data Processing}
    \begin{tabular}{@{}lccccccl@{}}
        \toprule
        \textbf{Category} & \textbf{Techniques} \\
        \midrule
        Statistical Methods & Mean, Median, Mode \\
        Distance-Based Methods & k-Nearest Neighbors (KNN)\cite{Murti2019-iu}, Combined (MICE + KNN) \\
        Regression-Based Methods & Multiple Imputation by Chained Equations (MICE) \cite{Azur2011-vz} \\
        Ensemble Methods & MissForest \cite{Stekhoven2012-rb}, PAIN \\
        Neural Network-Based & Autoencoder \cite{Li2024-sd}\\
        Custom Methods & combined (KNN-Ordinal + MICE-Continuous), Updated SICE \cite{Bhagat2024-cu} \\
        \bottomrule
    \end{tabular}
    \label{tab:Categorization of Techniques in Data Processing}
\end{table}

\vspace{10pt} 
\textbf{Evaluation Metrics}

The evaluation framework incorporates metrics to measure accuracy, computational efficiency, and statistical alignment between observed and imputed data (see Table \ref{tab:Imputation Evaluation Metrics and Purpose}).

\begin{table}[h]
    \centering
    \caption{Imputation Evaluation Metrics and Purpose}
    \begin{tabular}{@{}lcccccc@{}}
        \toprule
        \textbf{Metric} & \textbf{Purpose} \\
        \midrule
        Time Required & Measures computational time for imputation. \\
        NRMSE & Normalized error for continuous data. \\
        Sinkhorn Score & Compares observed and imputed distributions. \\
        MAE & Average absolute error for continuous data. \\
        PEV & Variance explained by imputed values. \\
        MMD & Tests similarity of data distributions. \\
        \bottomrule
    \end{tabular}
    \label{tab:Imputation Evaluation Metrics and Purpose}
\end{table}

Normalized root mean square error (NRMSE) and mean absolute error (MAE) assess imputation accuracy, while maximum mean discrepancy (MMD) and Sinkhorn Divergence evaluate statistical similarity. Predictive explained variance (PEV) ensures that imputed values preserve predictive power, and time measurements quantify computational feasibility. 

\medbreak
\textbf{Composite Imputer Framework }

The Composite Imputer operates through three sequential layers. The initial layer ie. baseline statistical layer implements a weighted combination of traditional (statistical methods) imputers with dynamic weighting: mean imputation $(w_1 = 0.3 \times (1 - \text{missingness\_ratio}))$, median imputation $(w_2 = 0.3 \times (1 - \text{missingness\_ratio}))$, and KNN imputation $(w_3 = 0.4 \times \text{missingness\_ratio})$. Simultaneously, k-Nearest Neighbors (k-NN) imputation is employed to address patterns within ordinal and mixed data types by leveraging the similarity among data points. The optimization of the k-value follows k = $\sqrt{n}$, where n represents the sample size. This weighted combination establishes a stabilizing foundation for subsequent imputation stages, with weights adjusting automatically based on the observed missingness patterns, with KNN gaining precedence as missingness increases.

\medbreak
The second layer ie. advanced imputation layer integrates ML and neural network approaches through two primary components. A Random Forest imputer operates with 100 decision trees, maximum depth of 10, minimum samples per leaf of 5, and feature subset size of $\sqrt{d}$ ($d$ = feature count) optimized for capturing non-linear relationships in mixed data types. Complementing this, an autoencoder network employs a symmetric architecture $[d \rightarrow d/2 \rightarrow d/4 \rightarrow d/8 \rightarrow d/4 \rightarrow d/2 \rightarrow d]$ with batch normalization layers, bottleneck layer, dropout regularization, and ReLU activation functions for reconstructing missing values by learning the underlying patterns in the data. The network training implements 100 epochs with early stopping (patience = 10), batch size of min(32, n/10), and Adam optimizer (learning rate = 0.001).

\medbreak
The final layer, the refinement layer, applies statistical post-processing to ensure consistency and minimize artifacts introduced during earlier stages. Outlier detection is conducted using thresholds based on 1.5 times the interquartile range (IQR), while Winsorization is applied at the 5th and 95th percentiles for continuous variables. For discrete variables, mode-based adjustments are implemented to align with observed distributions. Additionally, column-specific recalibration and distribution consistency checks are performed to further refine the imputed data. This layer mitigates noise and skew, ensuring the imputed dataset retains statistical integrity and aligns with the underlying data structure.

\section{Results}
Multiple imputation techniques are evaluated across datasets by calculating the mean performance metrics for each method and across different missingness percentages. The metrics includes NRMSE, Sinkhorn divergence, MAE, PEV, and MMD. To highlight the most effective methods, the best and second-best performers are identified for each dataset (refer Table \ref{tab:Performance Metrics of Best and Second Best Performers}). The "PAIN" emerged as the top-performing method across most datasets, excelling in NRMSE, Sinkhorn score, and MMD. Second-Best Model: MissForest is consistently the runner-up and is outperforming in specific cases, such as the Wine dataset.

\begin{table}[h!]
\centering
\renewcommand{\arraystretch}{1.2}
\caption{Performance Metrics of Best and Second Best Performers}
\label{tab:Performance Metrics of Best and Second Best Performers}
\adjustbox{max width=\textwidth}{%
\begin{tabular}{lccccccl}
    \hline
    \multicolumn{7}{c}{\textbf{Best Performer}} \\ \hline
    \textbf{Method} & \textbf{NRMSE} & \textbf{Sinkhorn Score} & \textbf{MAE Score} & \textbf{PEV Score} & \textbf{MMD Score} & \textbf{Dataset} \\ \hline
    PAIN           & 0.0121         & 0.1274                 & 0.7986             & 0.9947            & 1.7328            & Absenteeism       \\ \hline
    PAIN           & 0.0139         & 0.2356                 & 2.2836             & 0.9873            & 1.4994            & Forest Fire Dataset \\ \hline
    PAIN           & 0.0184         & 0.1160                 & 0.7909             & 0.9937            & 1.7824            & Health Data       \\ \hline
    PAIN           & 0.0147         & 0.0213                 & 4.6609             & 0.9970            & 1.8634            & Concrete          \\ \hline
    Missforest     & 0.0100         & 0.0158                 & 0.4189             & 0.9589            & 1.3208            & Wine              \\ \hline
    \multicolumn{7}{c}{\textbf{Second Best Performer}} \\ \hline
    \textbf{Method} & \textbf{NRMSE} & \textbf{Sinkhorn Score} & \textbf{MAE Score} & \textbf{PEV Score} & \textbf{MMD Score} & \textbf{Dataset} \\ \hline
    Missforest     & 0.0135         & 0.1208                 & 1.0099             & 0.9933            & 1.7395            & Absenteeism       \\ \hline
    Missforest     & 0.0172         & 0.1522                 & 2.9096             & 0.9809            & 1.5098            & Forest Fire Dataset \\ \hline
    Missforest     & 0.0187         & 0.0491                 & 0.8274             & 0.9937            & 1.7846            & Health Data       \\ \hline
    Missforest     & 0.0152         & 0.0389                 & 5.1977             & 0.9966            & 1.8648            & Concrete          \\ \hline
    PAIN           & 0.0107         & 0.0421                 & 0.4525             & 0.9530            & 1.3085            & Wine              \\ \hline
\end{tabular}%
}
\end{table}

\medbreak
The PAIN is consistently demonstrating superior performance, achieving the lowest NRMSE across all datasets except for Wine Quality, where MissForest marginally outperformed it. This highlights its robustness in accurately reconstructing missing data. Additionally, the PAIN’s lower Sinkhorn scores and MMD values underscore its ability to preserve original data distributions, which is essential for downstream ML tasks.

\medbreak
While the PAIN generally produces lower MAE than other methods, this advantage came at the cost of higher computational requirements comparing to traditional methods like MissForest. However, its adaptability to datasets with complex structures, such as Absenteeism and Forest Fire, makes it a suitable choice for high-dimensional and highly correlated features. Even as missingness levels increases, the PAIN maintains smaller errors, reflecting its reliability under challenging conditions.

\medbreak
The imputer's marginally lower Sinkhorn scores and consistently lower MMD values further emphasize its ability to replicate both feature-level and overall data distributions. This is especially critical for predictive modeling tasks requiring high-fidelity data reconstruction. Although the computational cost of the PAIN is higher, this trade-off is justified in applications where accuracy is paramount, such as medical research or other high-stakes domains.

\begin{figure}[h!]
    \centering
    \begin{subfigure}[b]{0.8\textwidth}
        \centering
        \includegraphics[width=\textwidth]{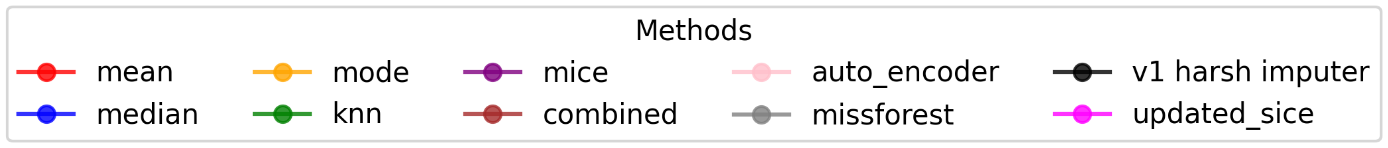}
    \end{subfigure}
    \begin{subfigure}[b]{0.45\textwidth}
        \centering
        \includegraphics[width=\textwidth]{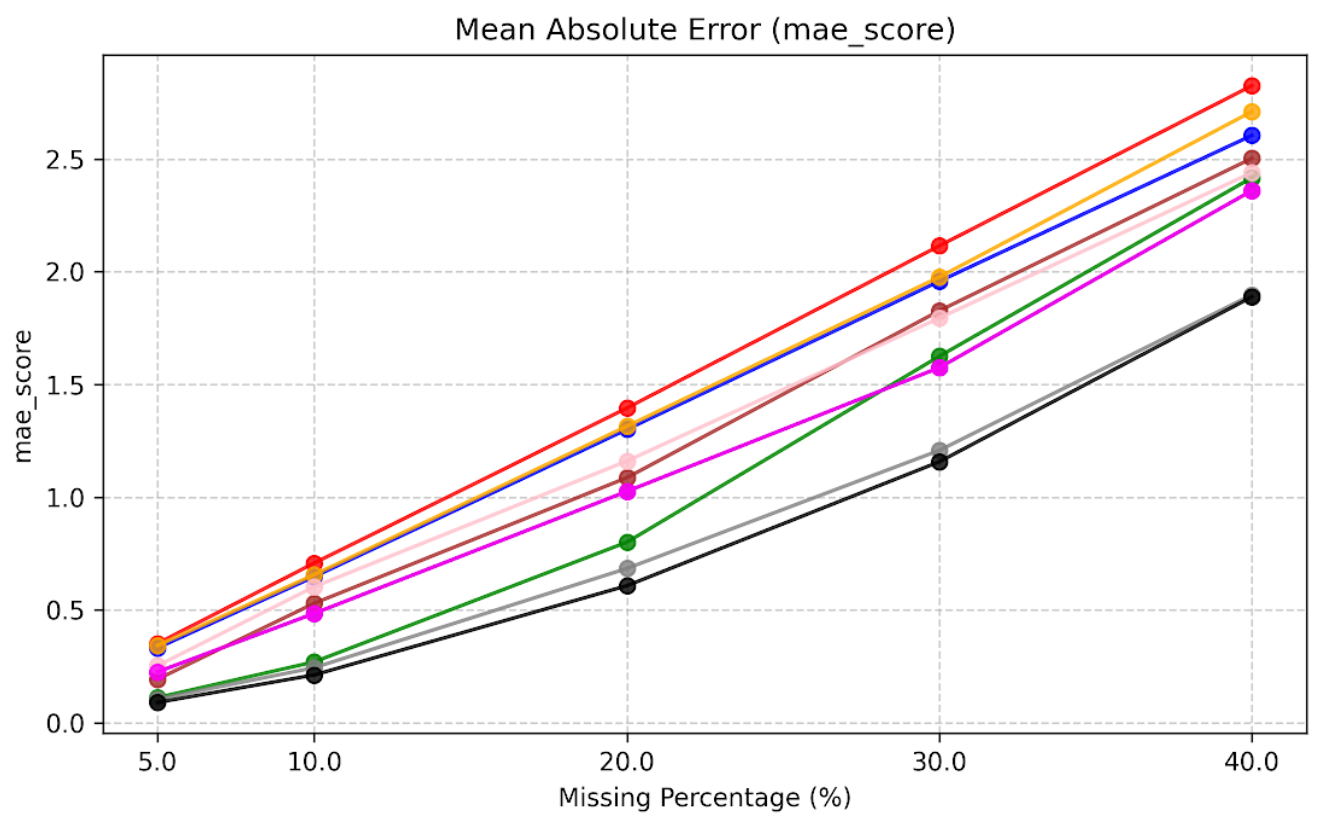}
    \end{subfigure}
    \hfill
    \begin{subfigure}[b]{0.45\textwidth}
        \centering
        \includegraphics[width=\textwidth]{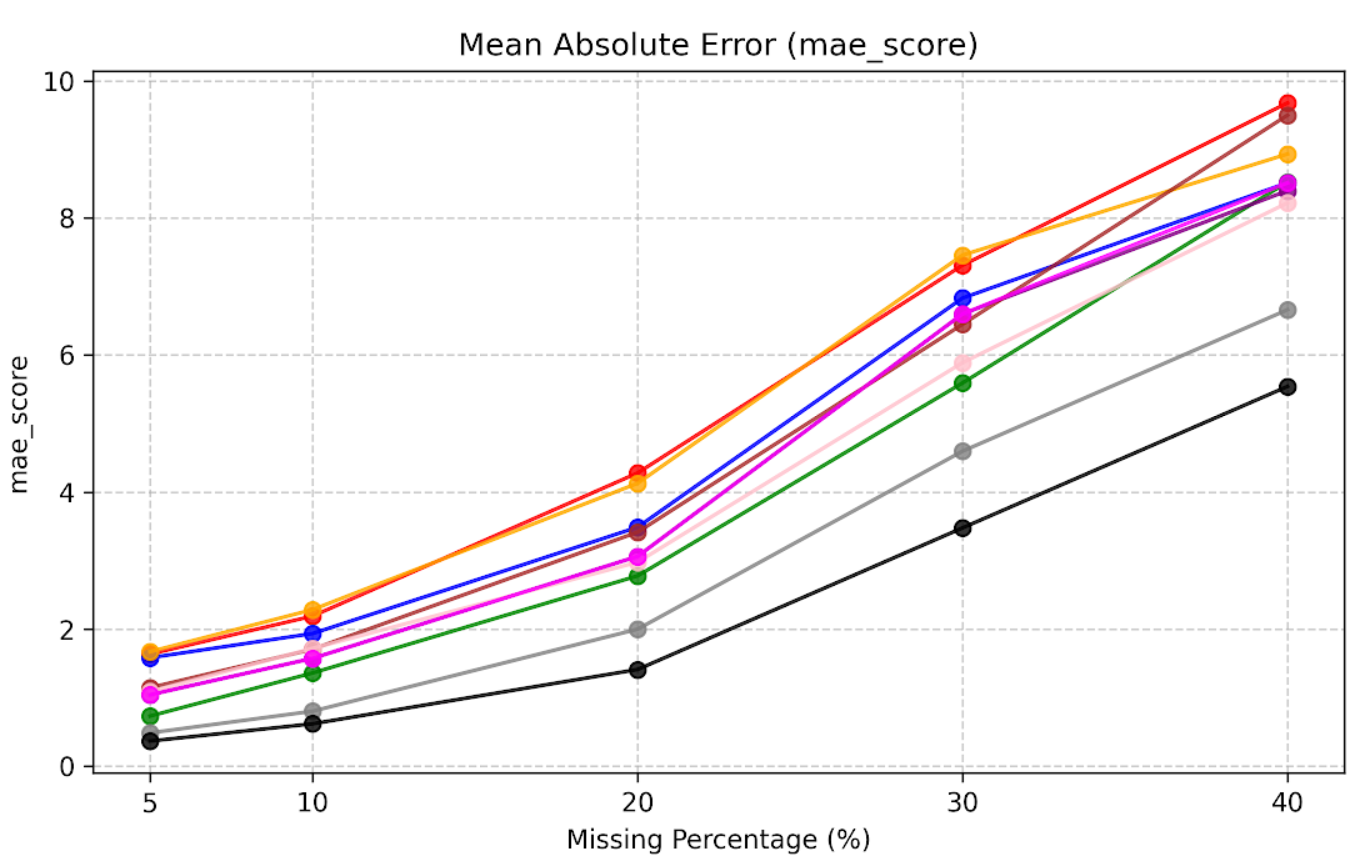}
    \end{subfigure}
    
    \caption{MAE of Maternal Health Risk Data and Forest Fire Data}
    \label{fig:MAE of Maternal Health Risk Data and Forest Fire Data}
\end{figure}

\medbreak
As observed in figure \ref{fig:MAE of Maternal Health Risk Data and Forest Fire Data}, the PAIN method (black line) consistently achieves the lowest MAE across all missingness levels, indicating its effectiveness in minimizing reconstruction errors. MissForest (gray line) follows closely, with slightly higher MAE values but still significantly outperforming simpler methods. As missingness increases from 5\% to 40\%, MAE rises for all methods, but PAIN and MissForest demonstrate slower growth, highlighting their robustness. The Forest Fire Data poses greater imputation challenges, as shown by higher MAE values compared to the Maternal Health Risk Data.

\begin{figure}[h!]
    \centering
    \begin{subfigure}[b]{0.45\textwidth}
        \centering
        \includegraphics[width=\textwidth]{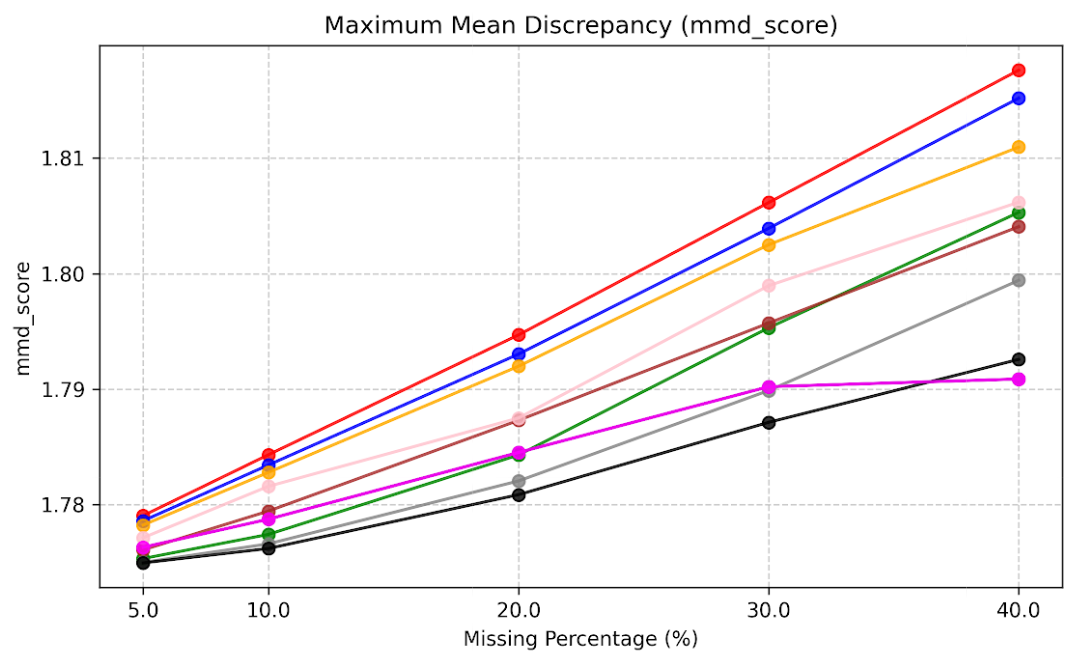}
    \end{subfigure}
    \hfill
    \begin{subfigure}[b]{0.45\textwidth}
        \centering
        \includegraphics[width=\textwidth]{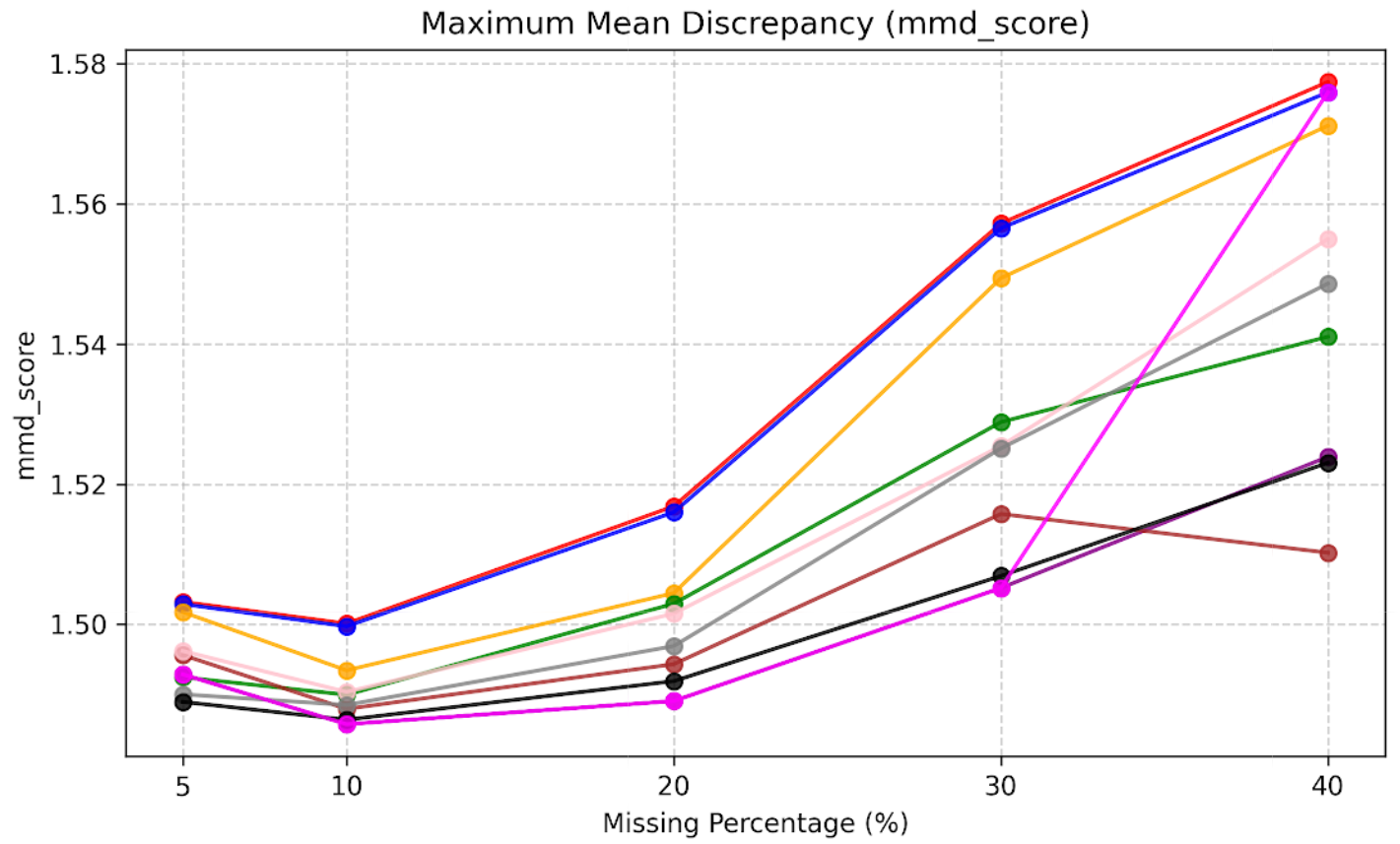}
    \end{subfigure}
    
    \caption{MMD of Maternal Health Risk Data and Forest Fire Data}
    \label{fig:MMD of Maternal Health Risk Data and Forest Fire Data}
\end{figure}

\medbreak
In figure \ref{fig:MMD of Maternal Health Risk Data and Forest Fire Data}, PAIN (black line) achieves the lowest MMD values, excelling in replicating original data distributions. MissForest (gray line) performs well but lags slightly behind PAIN. Advanced methods like the Combined (green) and Autoencoder (yellow) \cite{Pereira2020-vs} show moderate performance, while simpler methods such as mean and mode imputation fare poorly, particularly at higher missingness levels. The Forest Fire Data shows a more pronounced performance gap, emphasizing the challenges of maintaining data distribution accuracy under increased missingness.

\medbreak
Average performers such as MICE (purple line) and the Combined method (brown line) show moderate results, with a steadier increase in MAE compared to simpler methods. However, their performance gap relative to the best methods widens at higher missingness percentages. Poor performers, including mean (red line) and mode (orange line) imputation, exhibit the steepest MAE increases, particularly beyond 20\% missingness. This highlights their inability to handle substantial missing data effectively. The Forest Fire Data presents a greater challenge for all methods, as seen in the overall higher MAE values compared to the Maternal Health Risk Data (see figure \ref{fig:MAE of Maternal Health Risk Data and Forest Fire Data}, figure \ref{fig:MMD of Maternal Health Risk Data and Forest Fire Data}).

\medbreak
These findings emphasize the importance of selecting the imputation method based on dataset characteristics and the extent of missing data. Methods like PAIN and MissForest consistently showcase their ability to handle both low and high levels of missingness, making them reliable choices for complex imputation tasks.

\section{Discussion}
The performance of imputation techniques varies depending on the nature of data. Datasets with mixed data such as discrete and continuous variables pose unique challenges. Continuous variables benefit from methods that capture variation and magnitude, whereas discrete data require techniques that respect the category boundaries and avoid invalid intermediate values. Failure to address these differences can lead to biases and invalid imputations, reducing the reliability of the results, particularly for models that are sensitive to datatype consistency \cite{Joel2024-ri,May2023-jw}. Furthermore, data distribution significantly influences the imputation outcomes. For normally distributed data, simpler approaches, such as mean- or regression-based methods, work relatively well. However, non-normally distributed or skewed data complicates imputation, as standard methods often fail to capture asymmetries, leading to biased and inaccurate estimates \cite{Jadhav2019-ue,Austin2021-ai}.

\medbreak
This study evaluates the performance of various imputation techniques under diverse missingness levels and dataset characteristics. The missingness percentage is restricted deliberately to under 40\%, as datasets with over 50\% missing values introduce significant challenges. At high levels of missingness, the underlying patterns become sparse, leading to unreliable imputations, increased uncertainty, and a greater risk of bias. Such scenarios can substantially alter the original data distribution, thereby undermining the robustness of the downstream analyses \cite{Khattab2023-tr}. Metrics such as NRMSE, MAE, and MMD exhibit greater instability as missingness increases, with error scores rising sharply owing to the inability of imputations to accurately reconstruct the data \cite{Joel2024-ri,Jadhav2019-ue}.

\medbreak
Among the evaluated methods, advanced imputation techniques demonstrate superior performance. The PAIN is consistently achieving the lowest NRMSE, MAE, and MMD values, reflecting its ability to reconstruct data with high accuracy while preserving feature-level and overall distributions. This proves particularly effective for datasets with complex structures, such as Absenteeism and Forest Fire, characterizing by high-dimensional and correlating features. MissForest also performs well, notably outperforming the PAIN in preserving distributions in datasets like Wine. In contrast, simpler methods, including mean, mode, and median imputations, exhibits lower performance. These methods fail to account for underlying relationships, leading to distorted data distributions and reduced applicability in datasets with significant complexity or correlation.

\medbreak
In real-world imputation tasks, consistency across evaluation metrics is more critical than excelling in isolated measures. For instance, an approach that achieves low NRMSE but fails to preserve data distributions, as MMD or Sinkhorn divergence indicates high score, can compromise the integrity of downstream ML applications. Techniques such as the PAIN and MissForest are notable for maintaining balanced performance across metrics like NRMSE, MAE, MMD, and Sinkhorn scores, ensuring reliability across diverse scenarios.

\medbreak
DL techniques such as neural networks, GANs, and autoencoders have demonstrated significant potential in uncovering intricate data patterns and generating realistic imputations. However, their effectiveness comes with notable constraints. These methods often require extensive datasets—typically exceeding tens of thousands of samples—to avoid overfitting and ensure robust generalization. Furthermore, their deployment demands meticulous task-specific hyperparameter tuning, which can be both computationally intensive and time-consuming. For instance, training a GAN for missing data imputation on a moderately sized healthcare dataset might necessitate hours to days of optimization on high-performance GPUs, significantly inflating computational costs. This renders such methods less suitable for small datasets, such as those with 1,000–1,500 samples, commonly encountered in niche research domains. In contrast, simpler or hybrid techniques like MissForest or composite imputers strike a more practical balance by providing reliable imputations with considerably lower resource demands. 

\medbreak
While the PAIN demonstrates exceptional performance, it has higher computational costs comparing with traditional methods like MissForest. However, this trade-off is justified in scenarios prioritizing accuracy, such as medical research or high-stakes predictive modeling. Overall, the findings underscore the importance of selecting imputation techniques tailored to dataset complexity, missing data characteristics, and the requirements of downstream applications.

\medbreak
The study's findings extend beyond methodological comparisons, highlighting the critical need for context-aware imputation strategies. No universal approach can effectively address all data reconstruction challenges. Researchers must carefully consider dataset complexity, missing data characteristics, and the specific requirements of downstream applications. The varying performance across different evaluation metrics—including NMRSE, MAE, and MMD—emphasizes the importance of holistic assessment rather than relying on isolating performance indicators.

\medbreak
The broader implications of this research extend to multiple domains where data quality is paramount. In medical research, predictive modeling, and other high-stakes applications, the ability to accurately reconstruct missing data can significantly impact analytical outcomes. By understanding the nuanced limitations of current imputation techniques, researchers can develop more advanced approaches that balance computational efficiency with analytical precision. Ultimately, this study contributes to a more nuanced understanding of missing data reconstruction, providing a critical framework for future methodological innovations in data science and ML.

\section{Conclusion}
In conclusion, this research comprehensively evaluates the performance of various imputation techniques across diverse datasets and missingness levels. The results demonstrate that advanced imputation methods, such as the PAIN and MissForest, consistently outperform simpler methods in terms of NRMSE, MAE, and MMD. The PAIN emerged as the top-performing method across most datasets, excelling in NRMSE, Sinkhorn score, and MMD. MissForest is consistently the runner-up, outperforming in specific cases.

\medbreak
The findings highlight the importance of selecting imputation techniques that is tailored to dataset’s complexity, missing data characteristics, and the requirements of downstream applications. While the PAIN provided exceptional performance, its higher computational cost is a drawback compared to traditional methods. However, this trade-off is acceptable in high-stakes domains like medical research, where accuracy outweighs speed.

\medbreak
The study's results underscore the need for a balanced approach to imputation, considering multiple evaluation metrics and dataset characteristics. The PAIN, which demonstrates the superior performance in terms of NRMSE, MAE, and MMD, offers a promising solution for datasets with complex structures and high-dimensional features. Future research should focus on developing hybrid methods that combine the strengths of different imputation techniques, enabling more accurate and efficient data reconstruction.

\bibliography{IR.bib}
\bibliographystyle{unsrt}

\end{document}